%% file: RIFT_arxiv.tex
\newif\ifnamedversion
\namedversiontrue

\documentclass[letterpaper]{article}
\ifnamedversion
  \usepackage[preprint]{aaai2027}
\else
  \usepackage{aaai2027}
\fi
\usepackage{times}
\usepackage{helvet}
\usepackage{courier}
\usepackage[hyphens]{url}
\usepackage{natbib}
\usepackage{caption}
\usepackage{graphicx}
\usepackage{amsmath,amssymb}
\usepackage{algorithm}
\usepackage[noend]{algpseudocode}
\usepackage{subcaption}
\usepackage{cleveref}

\usepackage{xcolor}
\input{math_commands.tex}


\newcommand{\method}{\textsc{Rift}}
\newcommand{\XS}[1]{}

\def\XS#1{{\color{blue}{\bf [Xuesong:} {\it{#1}}{\bf ]}}}

\usepackage{newfloat}
\usepackage{listings}
\DeclareCaptionStyle{ruled}{labelfont=normalfont,labelsep=colon,strut=off}
\floatstyle{ruled}
\newfloat{listing}{tb}{lst}{}
\floatname{listing}{Listing}
\usepackage{booktabs}

\ifnamedversion
\else
\fi

\title{Keep the Future, Drop the Rollout: \method{} for World Action Models}
\ifnamedversion
  \author{\normalsize
    Chushan Zhang\textsuperscript{1,2}\quad
    Jinguang Tong\textsuperscript{1}\quad
    Xuesong Li\textsuperscript{1}\quad
    Yikai Wang\textsuperscript{3,*}\quad
    Hongdong Li\textsuperscript{1,*}}
  \affiliations{\small
    \textsuperscript{1}Australian National University\quad
    \textsuperscript{2}FreiNexus\quad
    \textsuperscript{3}Beijing Normal University\\
    \textsuperscript{*}Corresponding authors\\
    Project page: \url{https://github.com/ChushanZhang/RIFT}}
\else
  \author{Anonymous submission}
  \affiliations{}
\fi

\begin{document}

\maketitle

\begin{abstract}
\input{sections/0_abstract}
\end{abstract}

\input{sections/1_introduction}
\input{sections/2_related_work}
\input{sections/3_analysis}
\input{sections/4_method}
\input{sections/5_experiments}

\input{sections/6_conclusion}

\pagebreak
\bibliography{references}

\clearpage
\input{sections/A_appendix}

\end{document}

%% file: math_commands.tex
\usepackage{amsmath,amsfonts,bm}

\def\eqref#1{\cref{#1}}

\def\1{\bm{1}}

\DeclareMathAlphabet{\mathsfit}{\encodingdefault}{\sfdefault}{m}{sl}
\SetMathAlphabet{\mathsfit}{bold}{\encodingdefault}{\sfdefault}{bx}{n}

%% file: sections/0_abstract.tex
World action models (WAMs) condition robot actions on predicted futures, but iterative video rollout increases deployment latency.
We ask whether action generation requires the evolving rollout trajectory or only its future representation.
Across four WAMs on 40 simulated robotic manipulation tasks, paired closed-loop interventions show that blocking access to the future cache or reassigning its values changes execution and reduces success.
Yet in the evaluated co-denoising settings, reusing one fixed final-clean key/value (K/V) cache throughout action denoising nearly preserves unmodified execution, with $1.7$--$1.9$~cm end-effector average displacement error.
Obtaining this cache still requires iterative video generation.
We therefore propose \method{} (\emph{Rollout-free Imagination via Future Tokens}), which uses learned anticipation tokens to construct a complete future K/V cache in one backbone pass.
On LIBERO, \method{} achieves $98.8\%$ overall success, outperforming all evaluated rollout-based methods while yielding a \textbf{$\bm{3.1}$--$\bm{9.2\times}$} inference speedup.
Without further training, it achieves $81.1\%$ overall success on the out-of-distribution LIBERO-Plus benchmark, a $+9.7$ percentage-point improvement over the strongest evaluated baseline.
On RoboTwin, it achieves $92.9\%$ and $92.6\%$ success on clean and randomized scenes, respectively, the highest among the evaluated methods.
On real-world manipulation tasks, \method{} achieves $45.3\%$ average success,
a $+6.0$ percentage-point improvement over Fast-WAM-Joint.
These results support rollout-free future conditioning without iterative video generation at deployment.

%% file: sections/1_introduction.tex
\section{Introduction}
\label{sec:intro}

\begin{figure}[t]
    \centering
    \includegraphics[width=\columnwidth]{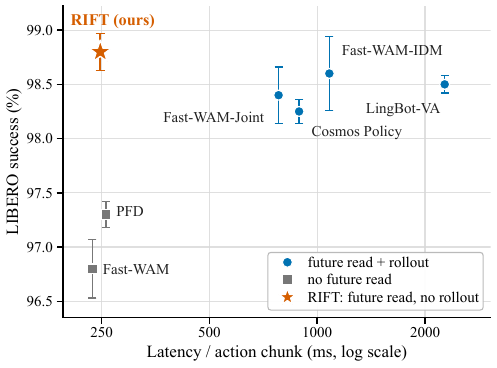}
    \caption{\textbf{Success rate vs. deployment latency.} Points show mean success rates, and
    error bars indicate $\pm 1$ standard deviation across three evaluation runs per method.
    Latency is measured in milliseconds per action chunk on one A800.
    \textcolor[HTML]{777777}{Grey squares} denote current-only methods;
    \textcolor[HTML]{0072B2}{blue circles} denote methods that read a future representation produced by rollout.
    \textcolor[HTML]{D55E00}{\method{}} achieves the highest success rate with latency close to the minimum.}
    \label{fig:teaser}
\end{figure}

World action models couple future video prediction with robot control. A single network predicts a short video and conditions its next actions on that prediction \citep{yuan2026fastwam,li2026causalworld,bi2025motus}. In matched settings, rollout-based policies achieve higher success than current-only variants. However, iterative video generation makes rollout-based systems incur $3.3\times$ to $9.6\times$ the latency of current-only deployment (\cref{fig:teaser}).

Fast-WAM drops the future branch after future-prediction co-training \citep{yuan2026fastwam}, whereas PFD distills a future-conditioned correction into the current-only path \citep{fang2026pfd}. Both reduce latency but retain a gap to policies that explicitly read a generated future. Yet the action expert consumes a future representation, whereas iterative rollout is the process that constructs it. Existing comparisons change both the availability of this representation and the process used to construct it, so their separate contributions remain unresolved. We therefore ask whether a WAM can preserve explicit test-time future conditioning without iterative video rollout.

To separate these factors, we examine the future cache, which contains the keys and values of future tokens at each layer. Joint co-denoising updates this cache at every denoising step, whereas the generate-then-act inverse dynamics model (IDM) uses a fixed final-clean cache computed from the fully denoised future video. Their similar success suggests that the representation, rather than its iterative trajectory, may provide the shared benefit. Success alone cannot reveal what the action reads.

We therefore intervene on this cache. The attention mask prevents video tokens from attending to action tokens, so the cache is an action-independent intervention site. We record it and replay action denoising with non-target inputs fixed, masking attention to future tokens, editing future values under the recorded keys, or replacing both keys and values with one fixed final-clean cache. Each closed-loop intervention uses the same initial state and policy seed as the unmodified policy. We measure success rate (SR) and end-effector average displacement error (EE-ADE), the average drift from the unmodified end-effector trajectory.

Across $2{,}000$ paired trials on all 40 LIBERO tasks, masking yields $18.7$~cm EE-ADE and reduces success from $98.4\%$ to $9.7\%$. Spatial permutation and temporal swapping yield similar EE-ADE ($14.3$ and $15.6$~cm) but sharply different success ($65.2\%$ and $0.7\%$), showing that the action reads future content at its assigned positions. In contrast, reusing one fixed final-clean K/V cache yields only $1.9$~cm EE-ADE and $97.9\%$ success. Actions depend on future content and its organization, yet one fixed cache nearly preserves execution.

This cache still requires iterative video generation. We therefore ask whether one forward pass can produce an effective, complete future cache. We propose \method{} (\emph{Rollout-free Imagination via Future Tokens}), which places learned anticipation tokens at future temporal positions and fills their per-layer K/V cache with one video-backbone pass. We shape the anticipation states with conditional flow matching, using a distributional objective rather than direct L2 regression to the single observed future. Deployment requires one cache prefill followed by the ordinary action flow, without video diffusion or video decoding at test time.

On LIBERO, \method{} achieves $98.8\%$ overall success with a $3.1$--$9.2\times$ inference speedup over the evaluated rollout-based baselines. Without further training, it achieves $81.1\%$ overall success on LIBERO-Plus, a $+9.7$ percentage-point improvement over the strongest evaluated baseline. On RoboTwin~2.0, it achieves the highest observed success rates among the evaluated methods in both clean and randomized scenes. On three real-world tasks, \method{} achieves $45.3\%$ average success, a $+6.0$ percentage-point improvement over Fast-WAM-Joint, with $60.7\%$ lower action-chunk latency.

We make three contributions.
\begin{enumerate}
    \item \textbf{We introduce a paired closed-loop intervention protocol} to measure how future caches affect executed trajectories and task success.
    \item \textbf{We show that actions are sensitive to future content and its spatial and temporal organization.} Yet one fixed final-clean K/V cache nearly preserves execution in the evaluated co-denoising settings.
    \item \textbf{We develop a rollout-free future interface.} \method{} constructs a complete future K/V cache in one backbone pass and achieves higher overall success than the evaluated baselines across three simulation benchmarks and three real-world tasks. Its latency remains close to current-only Fast-WAM on LIBERO and the real robot.
\end{enumerate}

%% file: sections/2_related_work.tex
\section{Related Work}
\label{sec:related}

\paragraph{Vision-language-action policies.}
Vision-language-action (VLA) policies produce robot actions from visual observations
and language instructions \citep{zitkovich2023rt2,kim2024openvla,black2024pi0,
physicalintelligence2025pi05,bjorck2025grootn1,shukor2025smolvla,gemini2025robotics}.
Many use diffusion or flow matching to refine a sequence of actions over several steps
\citep{chi2023diffusion,liu2024rdt1b,wen2025dexvla}.
These policies provide a useful efficiency reference because they do not also generate future video.
Our aim is to retain the benefits of future prediction at similarly low deployment latency.

\paragraph{World action models.}
World models predict how an environment changes to support planning and control
\citep{schrittwieser2020muzero,hafner2023dreamer,hansen2024tdmpc2}.
In robot manipulation, generated future images or videos can guide the choice of actions
\citep{du2023unipi,black2024imageediting,bharadhwaj2024gen2act,zhou2024robodreamer}.
Other methods learn video prediction and action generation together
\citep{cheang2024gr2,li2025uva,zhu2025unifiedworldmodels,cen2025rynnvla}.
Within this family, one design generates a future video before predicting actions, while another
refines video and actions together
\citep{ye2026worldactionmodels,yuan2026fastwam,li2026causalworld,bi2025motus,kim2026cosmospolicy}.
In the former, action generation can use a fixed future representation; in the latter, this
representation changes during generation. Both designs incur the cost of iterative video generation.
We test whether the action generator needs these successive updates or can reuse one fixed future
representation.

\paragraph{Learning from future observations.}
Several methods learn from future observations during training while avoiding video
generation at deployment. Fast-WAM trains video and action prediction together, but generates actions
from the current observation at inference \citep{yuan2026fastwam}.
PFD uses a teacher with access to true future observations to train a correction for a policy that
uses only current observations \citep{fang2026pfd,vapnik2009privileged,chen2019learningbycheating}.
Other methods retain learned future representations at deployment.
FLARE and DreamVLA train internal features to capture future observations or scene properties
\citep{zheng2025flare,zhang2025dreamvla}.
Being-H0.7 trains such features with a separate branch that sees future observations and is removed
at inference \citep{luo2026beingh07}.
EvoScene-VLA uses future observations to supervise a scene representation that is updated by actions
and carried across control steps \citep{zhang2026evoscenevla}.
\method{} also retains a future representation, but constructs it directly in the form used by the
WAM action generator: the key/value cache. One backbone pass fills this cache at every layer, and
the action generator reuses it throughout action generation.

\paragraph{Interventions for understanding model behavior.}
Probing methods train simple predictors to recover information from a model's
internal features
\citep{alain2016linearprobes,belinkov2022probing}.
This shows what information is available, but does not establish that the model uses it.
Attention weights alone are also insufficient evidence of use
\citep{jain2019attention,serrano2019attention}.
Causal tracing tests the effect of selected internal representations by changing them and measuring
the model's response \citep{meng2022locating}.
We apply this idea to future caches and measure changes in executed robot trajectories and task
success.

\paragraph{Runtime failure monitoring.}
\citet{rao2026geometry} examine how an action prediction changes during generation
to estimate uncertainty and detect potential failures.
Our optional monitor compares two estimates of the future and uses their disagreement as a
warning signal. This monitor is separate from the action policy and excluded from reported latency
measurements; \cref{sec:app_uncertainty_monitor} provides the details.

%% file: sections/3_analysis.tex
\section{The Role of Future Representations in Action Generation}
\label{sec:analysis}

\input{sections/figures_main}

\subsection{The channel we edit}
\label{sec:instrument}

With paired closed-loop interventions, we test whether actions depend on the future cache,
whether future values must stay at assigned positions, and whether the complete K/V cache must evolve
during denoising. Each model--intervention estimate uses $2{,}000$
paired trials across all $40$ LIBERO tasks. \Cref{fig:future_cache_execution} reports executed
EE-ADE and success rate.

Fast-WAM-Joint, Fast-WAM-IDM, Cosmos Policy, and LingBot-VA construct futures
differently but all provide per-layer keys and values associated with future tokens. We call these the
\emph{future cache}; \emph{Oracle} denotes the unmodified checkpoint used as the reference for interventions.

Because video tokens don't attend to action tokens, the cache is action-independent
given observation ($o$), language ($l$), and video-generation randomness. This property supports record and replay. The record
pass generates video normally and stores per-layer future K/V at every action-denoising step. Value
corruptions retain the recorded key trajectory and edit only the matched future values. The
final-clean control instead takes the final clean future from Oracle's iterative generation,
prefills its complete K/V once, and reuses that fixed cache at every action-denoising step. All
non-target inputs remain fixed. Exact replay of the unedited trajectory reproduces \emph{Oracle}. Shuffle and noise alter only future values, with recorded keys fixed.
We interpret these out-of-distribution perturbations alongside the frozen-present and final-clean K/V controls.

\subsection{Scoring each edit}

Action chunks are not directly comparable across architectures, so we execute paired
policies. For episode $i$, intervention $I$ and \emph{Oracle} share the initial state and policy seed. Let $\mathbf{x}^{I}_{i,t}$ and $\mathbf{x}^{O}_{i,t}$ denote their recorded end-effector positions after environment step $t$. EE-ADE averages their distance over the common executed prefix $T_i$:
\begin{equation}
\operatorname{EE\text{-}ADE}_{i}(I)
= \frac{1}{T_i}\sum_{t=1}^{T_i}
\left\lVert \mathbf{x}^{I}_{i,t}-\mathbf{x}^{O}_{i,t}\right\rVert_2 
\label{eq:ee_ade}
\end{equation}
We exclude the reset pose and use recorded simulator positions rather than integrating predicted actions. When runs end at different times, only their common prefix contributes. We first average EE-ADE over episodes within each task, then average the task means with equal weight. \Cref{sec:app_fig2_uncertainty} defines the bootstrap procedure that resamples tasks. EE-ADE measures drift from \emph{Oracle}; success measures task completion.
High EE-ADE with a similar success rate indicates altered end-effector motion at comparable task performance.
When the success rate also decreases, the motion deviation is accompanied by reduced task success.

\subsection{Intervention set}
\label{sec:idm_intervention}

The interventions map directly to the three questions. Masking blocks attention to future tokens;
norm-matched noise replaces future values; and frozen-present supplies plausible, non-predictive values.
Spatial shuffle permutes values within frames, temporal swap exchanges value frames, and final-clean
replay replaces the evolving complete cache with the same final-clean K/V at every action step. We apply each supported edit to Fast-WAM-Joint, Fast-WAM-IDM, Cosmos Policy, and LingBot-VA, comparing each with its Oracle reference.
Claims are therefore within-model; cross-model magnitudes remain descriptive because architectures
and checkpoints differ.

\subsection{Finding 1: WAM action experts use future values at their assigned positions}
\label{sec:finding1}

Across all four WAMs, masking yields $11.8$--$20.4$~cm EE-ADE and reduces success
from $98.4$--$98.6\%$ to $0.0$--$32.0\%$. Under recorded keys, noise yields
$10.5$--$17.9$~cm and at most $40.9\%$ success, while frozen-present values yield
$18.1$--$21.3$~cm and at most $6.5\%$. These within-model effects show that action experts use
meaningful future values.

Position also matters: spatial shuffle yields $5.0$--$19.8$~cm and
$0.0$--$84.5\%$ success across all four, while temporal swap yields $15.6$--$16.3$~cm and
$0.0$--$69.0\%$ on Joint, IDM, and LingBot-VA. Every supported edit changes execution and lowers
own-model success, so future values are not an unordered pool. Severity differs by interface:
temporal swap is worse for Joint and IDM, spatial shuffle for LingBot-VA, and Cosmos-2 exposes no
temporal swap. Thus, position sensitivity is shared, not a universal severity ordering.

\subsection{Finding 2: One fixed final-clean K/V cache nearly preserves execution}
\label{sec:finding2}

Finding~1 establishes content and position sensitivity, not whether the complete
cache must evolve. Where supported, final-clean replay replaces the entire future-cache trajectory
with one final-clean cache, holding both K and V fixed at every action-denoising step.

For Joint and Cosmos-2, final-clean K/V replay gives $1.9/1.7$~cm EE-ADE and
$97.9/98.2\%$ success; the corresponding Oracle policies reach $98.4/98.4\%$. These are the smallest nonzero EE-ADEs
among compatible edits. IDM and LingBot-VA already expose one fixed final-clean K/V cache, so this
replay is identical by construction and structurally N/A.

The fixed cache nearly preserves Oracle execution in Joint and Cosmos-2,
but still requires iterative video generation.
The full-cache intervention also does not isolate the separate
contributions of keys and values. \Cref{sec:method} tests whether a single forward pass can construct
the complete future K/V cache used by the action expert.

%% file: sections/figures_main.tex
\begin{figure}[t]
    \centering
    \includegraphics[width=\columnwidth]{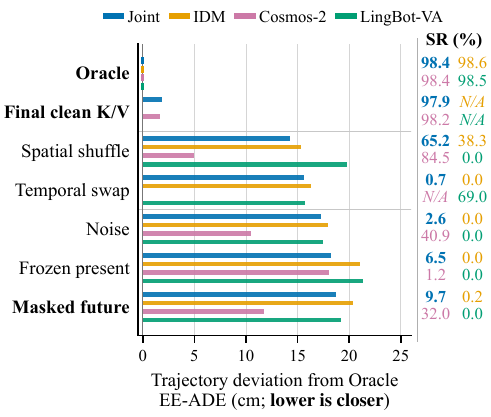}
    \caption{\textbf{A fixed final-clean K/V cache nearly preserves Oracle execution.}
    Masking attention to future tokens or perturbing future values alters trajectories and reduces success across four WAMs.
    For Joint and Cosmos-2, one final-clean K/V cache obtained from rollout is reused throughout action denoising, yielding trajectories and success rates close to Oracle.
    Oracle denotes each model's unmodified policy.
    Bars show end-effector trajectory deviation from Oracle, with success rates reported on the right.
    N/A marks structurally redundant or unsupported interventions.
    }
    \label{fig:future_cache_execution}
\end{figure}

%% file: sections/4_method.tex
\section{\method{}: One-pass Future-Token Imagination}
\label{sec:method}

\subsection{From findings to our design}

A fixed final-clean K/V cache nearly preserves Oracle execution in Joint and Cosmos-2,
but still requires iterative video generation. \method{} tests whether one forward pass can produce an effective future cache.

\begin{figure*}[t]
  \centering
    \includegraphics[width=1.0\textwidth]{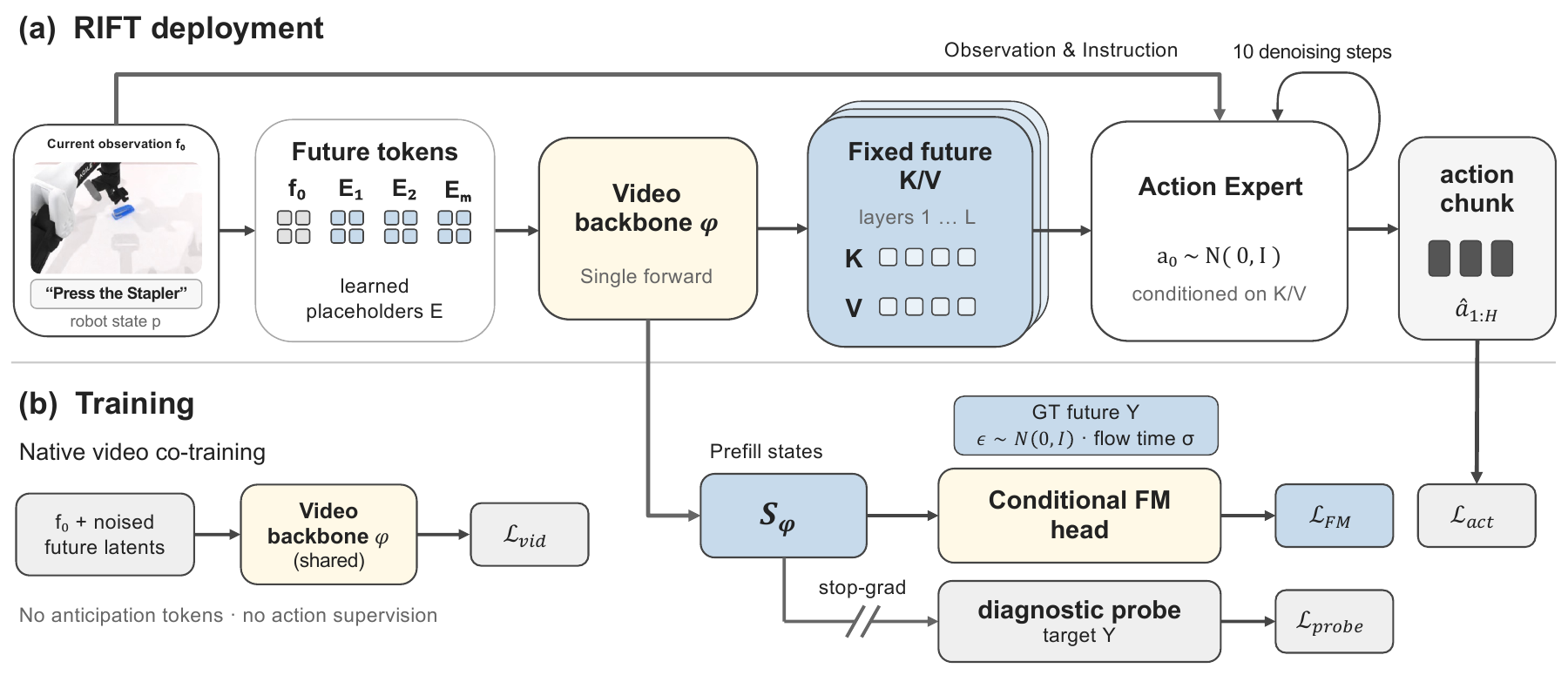}
  \caption{\textbf{\method{} training and deployment.}
  \textbf{(a)} At policy-only deployment, one backbone pass maps first-frame and anticipation tokens
  to a fixed future K/V cache at each layer, reused throughout action denoising.
  \textbf{(b)} Training combines native video supervision with action, conditional-FM, and
  stopped-gradient MSE probe losses from a deployment-matched pass.}
  \label{fig:rift_pipeline}
\end{figure*}

\subsection{Writing the cache in one pass}
\label{sec:method_arch}

\method{} keeps Fast-WAM-Joint's architecture. For a
video stack with hidden width $d$ and $L$ layers, it replaces rolled-out future tokens with learned
anticipation tokens $E\in\mathbb{R}^{m\times d}$. Each token
inherits its corresponding future spatiotemporal index. If each latent frame contains $n$ tokens and
the clip contains $T_{\mathrm{lat}}$ latent frames, full alignment uses
$m=n(T_{\mathrm{lat}}-1)$. As \cref{fig:rift_pipeline} shows, we use full alignment with $m=196$ on
LIBERO and $m=240$ on RoboTwin~2.0.

Let $f_0(o)$ denote the first-frame tokens extracted from observation $o$; the remaining observation
and instruction $l$ enter through the backbone's original conditioning path.
Let $\phi$ collect the shared video expert's parameters and the learned tokens $E$.
A single forward pass through the video expert computes the keys and values of all anticipation tokens:

\begin{equation}
\begin{split}
    \mathcal{C}_\phi(o,l)
    &= \left\{\left(K_E^{(\ell)},V_E^{(\ell)}\right)\right\}_{\ell=1}^{L} \\
    &= \mathrm{CachePrefill}_\phi\!\left([f_0(o);E],o,l\right)
\end{split}
\label{eq:future_cache}
\end{equation}
where
$K_E^{(\ell)},V_E^{(\ell)}\in\mathbb{R}^{m\times d}$ denote the layer-$\ell$ keys and values of the
$m$ anticipation tokens.
We retain Fast-WAM-Joint's mask: first-frame tokens attend only the observed frame,
anticipation tokens attend the first frame and one another, and video tokens never attend action
tokens. The cache is therefore action-independent.

The action expert attends to $\mathcal C_\phi$ at each layer using the original attention mechanism:
\begin{equation}
    \hat a_{1:H}=\mathrm{ActionDenoise}\left(o,l;\mathcal C_\phi(o,l)\right)
\label{eq:read}
\end{equation}
Here, $H$ is the action-chunk horizon; we use $H=32$. $\mathcal C_\phi$ has no
action-flow index: the same K/V serve every denoising
evaluation, as in Finding~2. Deployment requires one prefill
per chunk, with no rollout or VAE decoding.

\subsection{Training}
\label{sec:method_train}

Training uses two forwards through the same video expert per optimization step. Deployment retains
only the second forward's cache prefill.

\paragraph{Native video supervision.}
The first forward applies native video-flow loss $\mathcal L_{\mathrm{vid}}$
\citep{yuan2026fastwam} to a clean-first-frame clip with noised future latents and no anticipation
tokens or action supervision. This preserves dynamics supervision while changing only the deployment
interface.

\paragraph{Deployment-matched action training.}
The second forward matches deployment through input $[f_0;E]$ and its attention mask.
We train the action expert on this cache using unperturbed first-frame inputs
and Fast-WAM's action flow-matching loss $\mathcal L_{\mathrm{act}}$ \citep{yuan2026fastwam}.
After $70\%$ of the configured curriculum horizon, the probability and scale of first-frame latent
noise rise linearly from zero to $0.3$ and $0.06$ times the latent standard deviation. Examples with perturbed first-frame inputs
retain $\mathcal L_{\mathrm{FM}}$ and $\mathcal L_{\mathrm{probe}}$ but are excluded from the action loss.
Thus, $\mathcal L_{\mathrm{act}}$ is averaged only over unperturbed examples.

\paragraph{Conditional flow-matching supervision.}
Let $S_\phi\in\mathbb{R}^{m\times d}$ be the final anticipation states from the same
deployment-matched forward and $Y\in\mathbb{R}^{m\times d_y}$ the aligned ground-truth future latent
patches, where $d_y$ is the dimension of one flattened latent patch. Under a multimodal conditional
distribution, direct $\ell_2$ regression has a conditional-mean optimum and can average distinct valid
futures; we instead use conditional flow matching \citep{lipman2023flow} as a distributional auxiliary
objective. We sample
$\epsilon\sim\mathcal N(0,I)$ and $\sigma\in[0,1]$ with the native video-flow schedule, then define
\begin{equation}
    X_\sigma=(1-\sigma)Y+\sigma\epsilon,
    \qquad v_\sigma^\star=\epsilon-Y
\label{eq:fm_path}
\end{equation}
Conditioned on $S_\phi$, the training-only FM head $v_\psi$, parameterized by $\psi$,
predicts the velocity from $(X_\sigma,\sigma)$. Using the native video-flow timestep weight
$w_{\mathrm{vid}}(\sigma)$, its loss is
\begin{equation}
    \mathcal L_{\mathrm{FM}}
    =\mathbb E_{Y,\epsilon,\sigma}\!\left[
    w_{\mathrm{vid}}(\sigma)
    \left\|v_\psi(X_\sigma,\sigma;S_\phi)-v_\sigma^\star\right\|_2^2
    \right]
\label{eq:fm_loss}
\end{equation}
This auxiliary head supervises the video expert and anticipation tokens during training but enters neither
action nor policy-only deployment.

\paragraph{Stopped-gradient linear probe.}
The L2 variant provides a deterministic future-latent readout. In the final
model, we retain this view as a linear probe,
$\hat Y_{\mathrm{L2}}=g_\omega(\operatorname{RMS}(\operatorname{stopgrad}(S_\phi)))$.
  We train it with MSE,
  \[
  \mathcal L_{\mathrm{probe}}
  =\operatorname{mean}\!\left((\hat Y_{\mathrm{L2}}-Y)^2\right).
  \]
  The reduction weights every future-token and latent-channel squared residual uniformly before the batch mean.
Detached input confines this loss to the probe; \cref{sec:app_training_details} details the reduction.
For optional monitoring, the probe point prediction and conditional-FM
samples define a normalized disagreement score. Both read the same anticipation states, but neither
feeds the controller; \cref{sec:app_uncertainty_monitor} defines the score.

\paragraph{Objective and gradient routes.}
Both forwards share the video expert. $\mathcal L_{\mathrm{vid}}$ updates this expert
and its video head; $\mathcal L_{\mathrm{act}}$ updates the action expert and backpropagates through
the cache into the shared video expert and $E$; $\mathcal L_{\mathrm{FM}}$ updates the shared video
expert, $E$, and the FM head; and detached $\mathcal L_{\mathrm{probe}}$ updates only the probe.
\begin{equation}
    \mathcal L=\mathcal L_{\mathrm{vid}}+\mathcal L_{\mathrm{act}}
    +\lambda_{\mathrm{FM}}\mathcal L_{\mathrm{FM}}
    +\lambda_{\mathrm{probe}}\mathcal L_{\mathrm{probe}},
\label{eq:objective}
\end{equation}
Both auxiliary weights stay at $1$ for the first $70\%$ of the curriculum horizon, then follow a
  cosine decay to $0.2$ over the final $30\%$. Policy-only deployment retains one $[f_0;E]$ prefill, fixed
$\mathcal C_\phi$, and the standard action flow; \cref{sec:app_training_details} gives remaining
settings.

%% file: sections/5_experiments.tex
\section{Experiments}
\label{sec:experiments}

\subsection{Setup}
\label{sec:exp_setup}

\paragraph{Implementation.}
For LIBERO, Fast-WAM, Fast-WAM-Joint, Fast-WAM-IDM, and \method{} share the same
Wan2.2-5B \citep{wan2025} Fast-WAM backbone, training data, and $20$k-step budget, so differences
come from the future interface rather than scale or data. We additionally compare with the
Fast-WAM-based released PFD checkpoint \citep{fang2026pfd} and the embodied-pretrained LingBot-VA
\citep{li2026causalworld}.
Unless stated otherwise, \method{} uses conditional flow matching, while \method{}-L2 uses
direct L2 regression. Latency is milliseconds per
action chunk on one A800 under each method's own denoising configuration; \method{}'s figure covers
its cache prefill and action denoising and excludes optional diagnostic readouts. In the tables,
\emph{Fut.\ attn.} denotes attention to future tokens; \emph{rollout} denotes iterative
future generation at deployment.

\paragraph{Benchmarks.}
LIBERO \citep{liu2023libero} has $40$ tasks across its Spatial, Object, Goal, and Long suites. We
evaluate each method in three runs of $2{,}000$ trials ($50$ episodes per task) and report the
mean and standard deviation across runs. For
out-of-distribution (OOD) evaluation on LIBERO-Plus \citep{fei2026liberoplus}, we run Fast-WAM,
Fast-WAM-Joint, Fast-WAM-IDM, and \method{} with one rollout on each of the benchmark's $10{,}030$ variants, without further training. RoboTwin~2.0
\citep{chen2025robotwin} has $50$ bimanual tasks under clean and domain-randomized scenes. Following
\citet{yuan2026fastwam}, the matched Fast-WAM/Joint/PFD/\method{} family trains on $2{,}500$ clean-scene
and $25{,}000$ randomized demonstrations for $30$k steps, with externally pretrained LingBot-VA for
comparison. Each checkpoint is evaluated for $100$ trials per task in each setting.
\subsection{Quantitative results}
\label{sec:exp_main}

\paragraph{LIBERO.}
In \Cref{tab:main}, \method{} achieves $98.8\%$ mean overall success, outperforming
rollout-based Joint, IDM, and LingBot-VA at $98.4\%$ to $98.6\%$.
Unlike these methods, \method{} requires only $247.9$ ms per action chunk,
reducing latency by $68.2\%$ to $89.1\%$ while remaining close to current-only
Fast-WAM at $235.7$ ms. Compared with rollout-free Fast-WAM and PFD, \method{}
improves success by $2.0$ and $1.5$ percentage points, respectively, at
comparable latency.

\input{tables/libero_main_results}

\begin{figure}[t]
  \centering
  \includegraphics[width=\columnwidth]{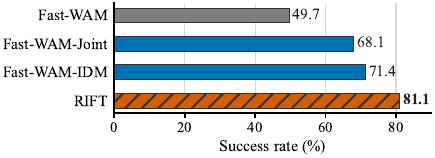}
  \caption{\textbf{Overall LIBERO-Plus OOD robustness.}
  Success across all $10{,}030$ variants, with one rollout per variant and no further training;
  bars are point estimates.}
  \label{fig:libero_plus_ood_overall}
\end{figure}

\paragraph{LIBERO-Plus OOD.}
Across four checkpoints and $10{,}030$ variants, \method{} achieves the highest overall success rate of $81.1\%$, a $+9.7$ percentage-point gain over
Fast-WAM-IDM (\cref{fig:libero_plus_ood_overall}). This shows robustness to OOD perturbations. See \cref{tab:libero_plus_ood} for the full breakdown.

\paragraph{RoboTwin~2.0.}
\label{sec:exp_robotwin}
The interface transfers to a second embodiment (\cref{tab:robotwin}; per-task rates in
\cref{tab:robotwin_pertask}). \method{} reaches $92.9/92.6$
on clean/randomized scenes, the best observed among the evaluated methods, against $92.5/92.1$ for
PFD, $92.4/91.4$ for rollout-based LingBot-VA, $91.9/91.6$ for Fast-WAM, and $91.0/91.1$ for
rollout-based Fast-WAM-Joint.
We observe higher overall success rates on RoboTwin~2.0 while retaining the
one-pass deployment path.

\input{tables/robotwin_main_results}

\subsection{Ablations}
\label{sec:exp_ablation}

\paragraph{Anticipation-token supervision.}
The variant \method{}-L2 regresses future latents with a direct L2 loss and reaches
$98.37{\scriptstyle\,\pm0.12}$; \method{} reaches $98.8{\scriptstyle\,\pm0.17}$. Both
use the same one-pass graph and $247.9$~ms cost, isolating supervision without deployment overhead.
The observed gain is $0.4$ percentage points; both variants use the same
training curriculum. \Cref{tab:libero_persuite} gives per-suite rates.

\begin{figure}[ht]
  \centering
  \includegraphics[width=\columnwidth]{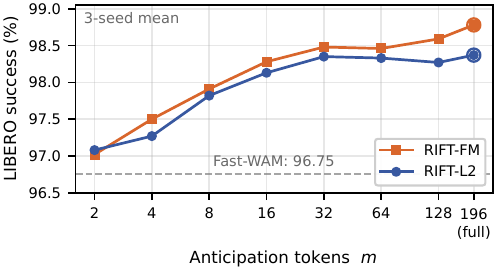}
  \caption{\textbf{Anticipation-interface capacity on LIBERO.}
  Mean success over four suites (three seeds) as token count $m$ grows for L2 and
  conditional-FM supervision. The dashed line is Fast-WAM without anticipation; full alignment
  ($m=196$) gives both model variants their best mean.}
  \label{fig:anticip_tokencount}
\end{figure}

\paragraph{The number of anticipation tokens.}
\label{sec:exp_tokencount} \Cref{fig:anticip_tokencount} sweeps $m=2$ to full alignment
($m=196$); current-only Fast-WAM is the no-cache $m=0$ reference ($96.75\%$). \method{}-L2 rises from
$97.08\%$ to $98.37\%$; conditional FM exceeds it from $m=4$ and peaks at $98.78\%$. Even small
interfaces beat the reference, and full alignment is best for both.

\input{sections/5_real_world}

\subsection{Visualizations and failure monitoring}

\paragraph{Real-world task executions.}
\Cref{fig:real_world_stack} illustrates \method{}'s autonomous execution
of the three real-world tasks. \Cref{fig:real_world_sequences} in the appendix shows
more detailed sequences.

\begin{figure}[t]
  \centering
  \includegraphics[width=\columnwidth,trim=7.2bp 3.89bp 0bp 2bp,clip]{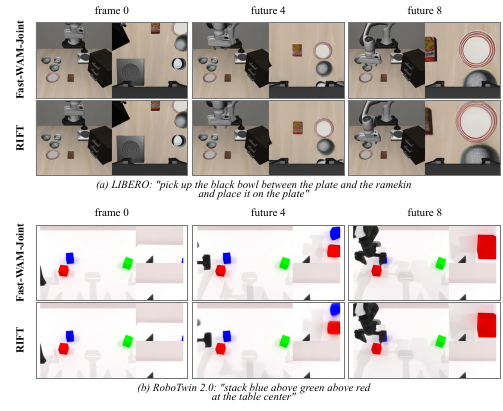}
  \caption{\textbf{Matched imagined futures.}
  From matched initial states, Fast-WAM-Joint iterative rollout (top) and \method{} one-pass
  anticipation decodes (bottom) show similar evolution at frames~0, 4, and~8 on LIBERO (upper block)
  and RoboTwin~2.0 (lower block). \method{}'s video decodes are for visualization only
and are omitted at deployment.}
  \label{fig:future_compare_main}
\end{figure}

\paragraph{Matched imagined futures.}
\Cref{fig:future_compare_main} compares Fast-WAM-Joint rollout and \method{} one-pass
decodes from matched starts on both benchmarks. Both evolve similarly at frames~0, 4, and~8. These
visuals diagnose future representations, not cache equivalence; \cref{sec:app_future_compare} shows
all frames.

\begin{figure}[!t]
  \centering
  \includegraphics[width=\columnwidth]{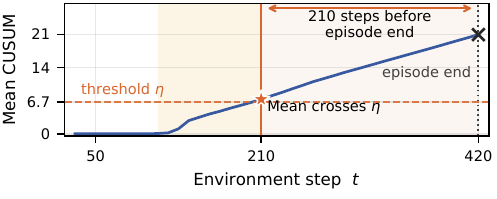}
  \caption{\textbf{L2--FM uncertainty warning over LIBERO.}
  Here $\eta$ is the CUSUM alarm threshold, conformally calibrated on
  the $1{,}967$ successful episodes; failures are excluded from calibration.
  The mean CUSUM curve over all failed evaluation rollouts exceeds $\eta$ at $t=210$,
  which is $210$ environment steps before the common episode endpoint at $t=420$.
  Individual alarm times can differ.
  Calibration and sampling details are given in \cref{sec:app_uncertainty_monitor}.}
  \label{fig:fm_probe_cusum}
\end{figure}

\paragraph{L2--FM uncertainty warning.}
The stopped-gradient L2 probe and conditional-FM head yield controller-independent
future estimates whose normalized disagreement defines a CUSUM warning
(\cref{sec:app_uncertainty_monitor}). We calibrate the alarm threshold $\eta$ on $1{,}967$
successful episodes. Across all failed evaluation rollouts, the mean CUSUM crosses the threshold
$\eta$ at $t=210$, which is $210$ environment steps before the common endpoint at $t=420$,
as shown in \cref{fig:fm_probe_cusum}.

%% file: tables/libero_main_results.tex
\begin{table}[t]
\centering
\caption{\textbf{LIBERO success and deployment cost.} SR is mean$\pm$std across three evaluation runs per method ($2{,}000$ trials per run). Latency is ms per action chunk on one A800 under each method's original denoising configuration. \emph{Emb.\ PT.}: embodied pretraining. $^{\dagger}$~released checkpoint evaluated by us; $^{*}$~our matched reproduction.}
\label{tab:main}
\setlength{\tabcolsep}{2.75pt}
\renewcommand{\arraystretch}{0.95}
\footnotesize
\begin{tabular}{@{}lccc r@{\hskip 6pt}r@{\hskip 6pt}r@{}}
\toprule
\textbf{Method} & \textbf{Emb.} & \textbf{Fut.} & \textbf{Roll-} & \textbf{SR} & \textbf{Lat.} & \textbf{Rel.} \\
                & \textbf{PT.}  & \textbf{attn.} & \textbf{out}  & (\%) & (ms) & \\
\midrule
Fast-WAM$^{\dagger}$      & $\times$   & $\times$   & $\times$   & $96.8{\scriptstyle\,\pm0.27}$ & \textbf{236}  & $1.0\times$ \\
PFD$^{\dagger}$           & $\times$   & $\times$   & $\times$   & $97.3{\scriptstyle\,\pm0.12}$ & 257  & $1.1\times$ \\
Fast-WAM-Joint$^{*}$      & $\times$   & \checkmark & \checkmark & $98.4{\scriptstyle\,\pm0.26}$ & 780  & $3.3\times$ \\
Fast-WAM-IDM$^{*}$        & $\times$   & \checkmark & \checkmark & $98.6{\scriptstyle\,\pm0.34}$ & 1081 & $4.6\times$ \\
LingBot-VA$^{\dagger}$    & \checkmark & \checkmark & \checkmark & $98.5{\scriptstyle\,\pm0.08}$ & 2270 & $9.6\times$ \\
\midrule
\textbf{\method{} (ours)} & $\times$   & \checkmark & $\times$   & $\mathbf{98.8}{\scriptstyle\,\pm0.17}$ & 248 & $1.1\times$ \\
\bottomrule
\end{tabular}
\end{table}

%% file: tables/robotwin_main_results.tex
\begin{table}[t]
\centering
\caption{\textbf{RoboTwin~2.0 closed-loop success} (\%) on clean and domain-randomized scenes. AVG is the mean of the success rates for the clean and randomized settings. \emph{Emb.\ PT.}: embodied pretraining.}
\label{tab:robotwin}
\setlength{\tabcolsep}{5pt}
\renewcommand{\arraystretch}{0.95}
\footnotesize
\begin{tabular}{@{}lccc r@{\hskip 6pt}r@{\hskip 6pt}r@{}}
\toprule
\textbf{Method} & \textbf{Emb.} & \textbf{Fut.} & \textbf{Roll-} & \textbf{Clean} & \textbf{Rand.} & \textbf{AVG} \\
                & \textbf{PT.}  & \textbf{attn.} & \textbf{out}  & & & \\
\midrule
LingBot-VA                & \checkmark & \checkmark & \checkmark & 92.4 & 91.4 & 91.9 \\
Fast-WAM-Joint            & $\times$   & \checkmark & \checkmark & 91.0 & 91.1 & 91.1 \\
Fast-WAM                  & $\times$   & $\times$   & $\times$   & 91.9 & 91.6 & 91.7 \\
PFD                       & $\times$   & $\times$   & $\times$   & 92.5 & 92.1 & 92.3 \\
\textbf{\method{} (ours)} & $\times$   & \checkmark & $\times$   & \textbf{92.9} & \textbf{92.6} & \textbf{92.8} \\
\bottomrule
\end{tabular}
\end{table}

%% file: sections/5_real_world.tex
\subsection{Real-world evaluation}
\label{sec:exp_real_world}

We evaluate \method{} against Fast-WAM and Fast-WAM-Joint on three
real-world manipulation tasks.

\paragraph{Experimental setup.}
We use a Galaxea R1 Lite robot with three RGB camera views as visual input.
We train all three methods for $10$ epochs on the same task-specific demonstrations from
RoboCOIN \citep{wu2025robocoin}.
Each model is evaluated on the three tasks in \cref{tab:real_world}, with
$50$ trials per task and $150$ trials in total per model.
We measure real-world inference latency on an NVIDIA RTX~4090 GPU.

\input{tables/real_world_results}

\paragraph{Experimental results.}
\Cref{tab:real_world} shows that \method{} achieves the highest observed success
rate on all three tasks. Its average success rate is $45.3\%$, compared with $39.3\%$ for
Fast-WAM-Joint and $26.7\%$ for Fast-WAM, a $+6.0$ percentage-point improvement over
the strongest evaluated baseline.
Its inference latency is $420$~ms per action chunk, compared with $1{,}069$~ms for
Fast-WAM-Joint and $404$~ms for Fast-WAM.

\begin{figure}[t]
  \centering
  \includegraphics[width=\columnwidth]{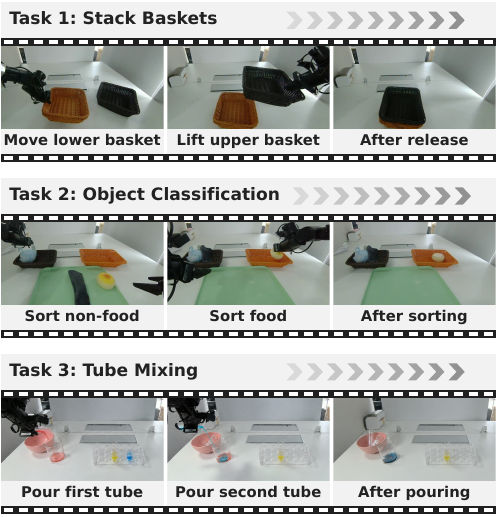}
  \caption{\textbf{Real-world task executions.}
  Key stages shown from the egocentric views, ordered from left to right.}
  \label{fig:real_world_stack}
\end{figure}

%% file: tables/real_world_results.tex
\begin{table}[t]
\centering
\caption{\textbf{Real-world evaluation.} Per-task entries report successes out
of $50$ trials. Overall success rate (SR) is computed over all $150$ trials per method.
Bold indicates the most successes, highest overall SR, or lowest latency.}
\label{tab:real_world}
\setlength{\tabcolsep}{3pt}
\renewcommand{\arraystretch}{1.08}
\small
\begin{tabular*}{\columnwidth}{@{\extracolsep{\fill}}lccc@{}}
\toprule
\textbf{Task} & \textbf{Fast-WAM} & \textbf{Fast-WAM-Joint} & \textbf{\method{}} \\
\midrule
Stack Baskets & 30/50 & 32/50 & \textbf{36}/50 \\
Object Classification & 10/50 & 19/50 & \textbf{22}/50 \\
Tube Mixing & 0/50 & 8/50 & \textbf{10}/50 \\
\midrule
\textbf{Overall SR (\%)} & 26.7 & 39.3 & \textbf{45.3} \\
Latency (ms/chunk) & \textbf{404} & 1069 & 420 \\
\bottomrule
\end{tabular*}
\end{table}

%% file: sections/6_conclusion.tex
\section{Conclusion}
\label{sec:conclusion}

Paired closed-loop interventions show that the evaluated WAMs depend on the content
and organization of their future caches. In Joint and Cosmos-2, one fixed final-clean cache nearly
preserves execution, although obtaining it still requires iterative video generation. \method{}
constructs a future K/V cache in one backbone pass and reuses it throughout action denoising.
Across the evaluated simulation benchmarks and three real-world tasks, it achieves higher observed
overall success than the evaluated baselines. On LIBERO and the real robot, its inference latency
remains close to current-only Fast-WAM. These results support retaining future conditioning without
iterative video generation at deployment.

%% file: sections/A_appendix.tex
\appendix

\section{Training implementation details}
\label{sec:app_training_details}

\paragraph{Loss normalization.}
For $\mathcal L_{\mathrm{vid}}$, we average unreduced squared error over channels and space, then
valid latent frames; for $\mathcal L_{\mathrm{FM}}$, over each patch vector, then valid future
patches. Both receive the native video-flow timestep weight per sample before the batch mean. For
$\mathcal L_{\mathrm{act}}$, we average over action dimensions, mask padding, average over the
horizon, and apply the base action-timestep weight before the batch mean. The coefficients of
$\mathcal L_{\mathrm{vid}}$ and $\mathcal L_{\mathrm{act}}$ are $1$.
$\lambda_{\mathrm{FM}}$ and $\lambda_{\mathrm{probe}}$ stay at $1$ for the first $70\%$ of the
curriculum, then cosine-decay to $0.2$; the stopped-gradient probe cannot affect the deployed
representation.

\paragraph{Deployment-matched perturbation.}
After $70\%$ of the curriculum, perturbation probability and standard deviation
rise linearly from $0$ to $0.3$ and from $0$ to $0.06$ times the latent standard deviation,
respectively. Training examples with perturbed first-frame inputs retain $\mathcal L_{\mathrm{FM}}$ and
$\mathcal L_{\mathrm{probe}}$ but are excluded from $\mathcal L_{\mathrm{act}}$.

\paragraph{Diagnostic probe.}
An RMS-normalized linear probe gives
$\hat Y_{\mathrm{L2}}=g_\omega(\operatorname{RMS}(\operatorname{stopgrad}(S_\phi)))$ with
$\mathcal L_{\mathrm{probe}}=\operatorname{mean}((\hat Y_{\mathrm{L2}}-Y)^2)$ over all batch
prediction elements; gradients reach only $g_\omega$.

\paragraph{Gradient routes.}
$\mathcal L_{\mathrm{vid}}$ updates the video expert and output head, but not the action expert;
$\mathcal L_{\mathrm{act}}$ updates the action expert and, through attended video states, the video
expert and anticipation tokens; $\mathcal L_{\mathrm{FM}}$ updates the video expert, anticipation
tokens, and FM head; detached $\mathcal L_{\mathrm{probe}}$ updates only the linear probe.

\paragraph{Architecture and training configuration.}
All in-house models share Wan2.2-5B's pretrained video DiT, text encoder, and video VAE
\citep{wan2025}. The action expert reuses the video branch at $d_a=1024$ ($1$B action; $6$B total),
with horizon $H=32$. Multi-camera images are concatenated before the VAE; video is temporally
downsampled $4\times$ to $9$ frames per chunk. Both branches use Fast-WAM's continuous flow
matching. We sample $u\sim\mathcal U[0,1)$, set $\sigma=5u/(1+4u)$ and $t=1000\sigma$, and reuse
the video scheduler for the FM head. Deployment uses $10$ action flow-matching steps,
classifier-free guidance $1.0$, and no video denoising or VAE decoding. Training uses AdamW at
$10^{-4}$ learning rate, $0.01$ weight decay, cosine annealing, mixed precision, and gradient
clipping at $1.0$, for $20$k LIBERO or $30$k RoboTwin~2.0 steps. Relative to Fast-WAM
\citep{yuan2026fastwam}, \method{} preserves these settings and adds only anticipation tokens,
$\mathcal L_{\mathrm{FM}}$, and the diagnostic probe.

\section{Uncertainty for the intervention study}
\label{sec:app_fig2_uncertainty}

\Cref{fig:future_cache_execution} reports point estimates; \cref{tab:fig2_eeade_ci} gives EE-ADE
intervals and the task-level resampling protocol.
\input{tables/fig2_eeade_ci_appendix}

\section{Per-suite LIBERO success rates}
\label{sec:app_libero_persuite}

\Cref{tab:libero_persuite} reports the per-suite results used in the main-text
ablation discussion.
\input{tables/libero_persuite_appendix}

\section{LIBERO-Plus OOD robustness}
\label{sec:app_libero_plus}

We directly evaluate LIBERO-trained checkpoints on the separate LIBERO-Plus benchmark without
further training. \method{} achieves $81.1\%$ overall success, outperforming Fast-WAM-IDM by
$9.7$ percentage points. It leads the four evaluated checkpoints across all seven
perturbation categories, all five labeled difficulty levels, and all four source suites;
\cref{tab:libero_plus_ood} reports the full breakdown.
\input{tables/libero_plus_ood_results}

\section{Optional L2--FM uncertainty warning}
\label{sec:app_uncertainty_monitor}

The final model retains direct L2's deterministic future-latent readout as a stopped-gradient
diagnostic probe. Both heads read the same anticipation states without feeding the
controller; this controller-independent comparison runs only as an optional monitor and is excluded
from policy-only deployment and reported latency. Let
$\mu_{\mathrm{L2}}=\hat Y_{\mathrm{L2}}\in\mathbb R^{m\times d_y}$ denote the probe estimate, let
$x_i^{\mathrm{FM}}\in\mathbb R^{m\times d_y}$ be the $i$th of $K$ samples from the conditional-FM
head, and let $\bar x_{\mathrm{FM}}=K^{-1}\sum_i x_i^{\mathrm{FM}}$ denote their sample mean at the
current replanning step. With $D=m d_y$ latent elements, we compute the discrepancy
using mean squared differences over those elements:
\begin{equation}
  d_{\mathrm{ratio}}
  =\frac{D^{-1}\lVert\mu_{\mathrm{L2}}-\bar x_{\mathrm{FM}}\rVert_F^2}
  {(KD)^{-1}\sum_{i=1}^{K}\lVert x_i^{\mathrm{FM}}-\bar x_{\mathrm{FM}}\rVert_F^2+\epsilon},
\label{eq:uncertainty_ratio}
\end{equation}
where $\epsilon=10^{-12}$ stabilizes the denominator. Large $d_{\mathrm{ratio}}$ means that the probe point
estimate departs from the FM mean beyond the dispersion of the FM samples. It is a warning statistic,
not a failure probability: both heads may still agree on the same wrong future. Unlike the
action-flow acceleration of \citet{rao2026geometry}, this score compares two estimators in
future-latent space.

\paragraph{FM sampling and monitoring frequency.}
At each replanning step, we generate $K=4$ samples as one batch, starting from standard Gaussian
noise with fixed seeds $0,1,2,3$. Each sample uses five explicit Euler steps on a time grid with
shift parameter $5.0$. The update is
$x_{j+1}=x_j+\Delta\sigma_j\,v_\psi(x_j,\tau_j;S_\phi)$, where $S_\phi$ denotes the final anticipation states
and $\tau_j$ is the FM time at integration step $j$.
After an initial wait of $30$ environment steps, the policy replans every $10$ environment steps
and executes $10$ actions from each predicted chunk. We compute one monitoring score per replan.
Let $n\geq1$ index these observations and $d_n=d_{\mathrm{ratio}}(n)$; observation $n$ is recorded
at environment step $t_n=30+10(n-1)$. The CUSUM updates once per observation:
\begin{equation}
  S_n=\max\!\left(0,S_{n-1}+d_n-\mu_d-0.25\sigma_d\right),
\label{eq:uncertainty_cusum}
\end{equation}
with $S_0=0$. An alarm occurs at the first $n$ for which $S_n>\eta$ and is assigned environment
step $t_n$.

\paragraph{Calibration.}
Of the $2{,}000$ evaluation episodes ($40$ LIBERO tasks with $50$ episodes each), we use the
$1{,}967$ successful episodes for calibration and exclude all $33$ failures. Pooling the
$29{,}059$ monitoring scores from successful episodes gives the arithmetic mean
$\mu_d=0.54082563$ and population standard deviation $\sigma_d=0.11140236$, computed with
divisor $29{,}059$. Thus, the reference level in
\cref{eq:uncertainty_cusum} is $\mu_d+0.25\sigma_d=0.56867622$.

Following the episode-peak CUSUM calibration rule of \citet{rao2026geometry}, we compute one
peak $P_i=\max_n S_n^{(i)}$ for each successful calibration episode. With $M=1{,}967$ and the
preselected level $\alpha=0.001$, the conformal order statistic is
\begin{equation}
  r=\left\lceil(M+1)(1-\alpha)\right\rceil=1{,}967,\qquad
  \eta=P_{(r)},
\label{eq:uncertainty_threshold}
\end{equation}
where $P_{(r)}$ is the $r$th smallest calibration peak. This selects the maximum peak,
$\eta=6.74138173$, displayed as $6.7$ in \cref{fig:fm_probe_cusum}. These successful episodes
are used for calibration and do not provide an independent estimate of the false-alarm rate.

\paragraph{Figure interpretation.}
\Cref{fig:fm_probe_cusum} averages $S_n$ over the $33$ failed rollouts at their corresponding
environment steps $t_n$. The mean curve first exceeds $\eta$ at $t=210$, which is $210$
environment steps before the common endpoint at $t=420$. This crossing is a property of the
mean curve; individual alarm times can differ.

\section{Real-world execution sequences}
\label{sec:app_real_world_sequences}

\paragraph{Task success criteria.}
Object Classification is successful only when all objects are placed in their
correct baskets. Success in Tube Mixing requires pouring the contents of both test tubes into
the same beaker and placing both tubes in their designated locations.

\begin{figure*}[t]
  \centering
  \includegraphics[width=\textwidth]{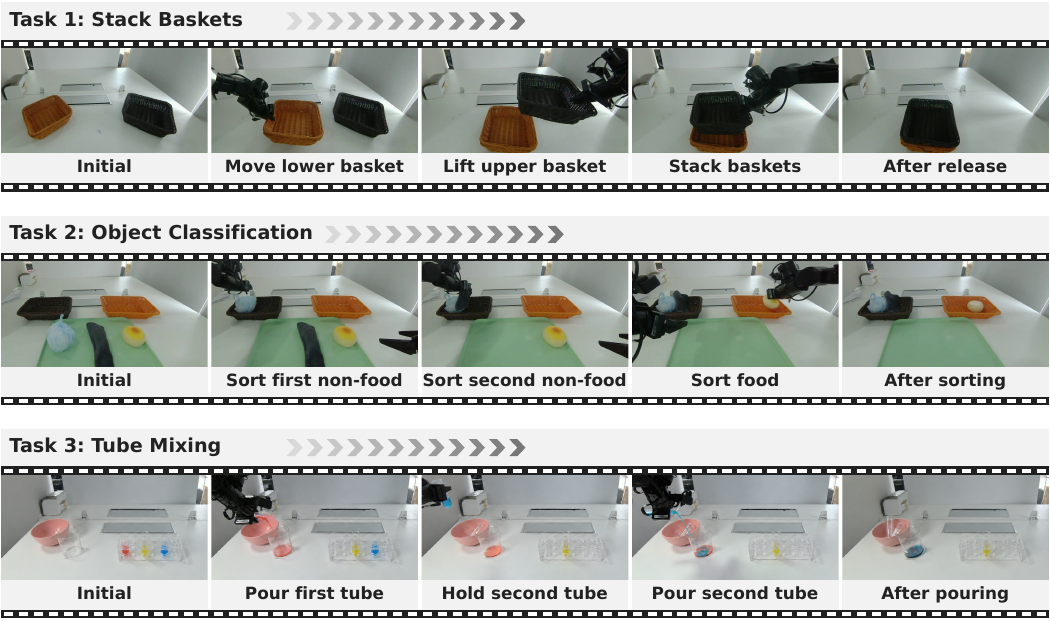}
  \caption{\textbf{Detailed real-world task executions.}
  Five successive stages per task, shown from the robot's viewpoint.
  Top: the robot repositions the lower basket, then lifts,
  stacks, and releases the upper basket. Middle: the robot places two non-food objects in the
  black basket and a food object in the orange basket. Bottom: the robot pours from the first
  test tube, picks up the second, and pours its contents into the same beaker. The final frame
  shows the resulting state after the arm withdraws.}
  \label{fig:real_world_sequences}
\end{figure*}

\Cref{fig:real_world_sequences} expands the three-frame summaries in
\cref{fig:real_world_stack} to five stages per task, all shown from the robot's viewpoint.
The main-paper panels are a subset of the images in each corresponding row.

\section{Limitations and future work}
\label{sec:app_limitations}

Evaluation on physical robots is limited to the Galaxea R1 Lite platform.
Transfer to other physical robot platforms and non-WAM fusion backbones remains future work.

\section{RoboTwin results and future decodes}
\label{sec:app_additional_results}
\label{sec:app_robotwin_pertask}
\label{sec:app_future_compare}

\begin{figure*}[ht]
    \centering
    \includegraphics[width=\textwidth,trim=12.5bp 4bp 2.4bp 3.8bp,clip]{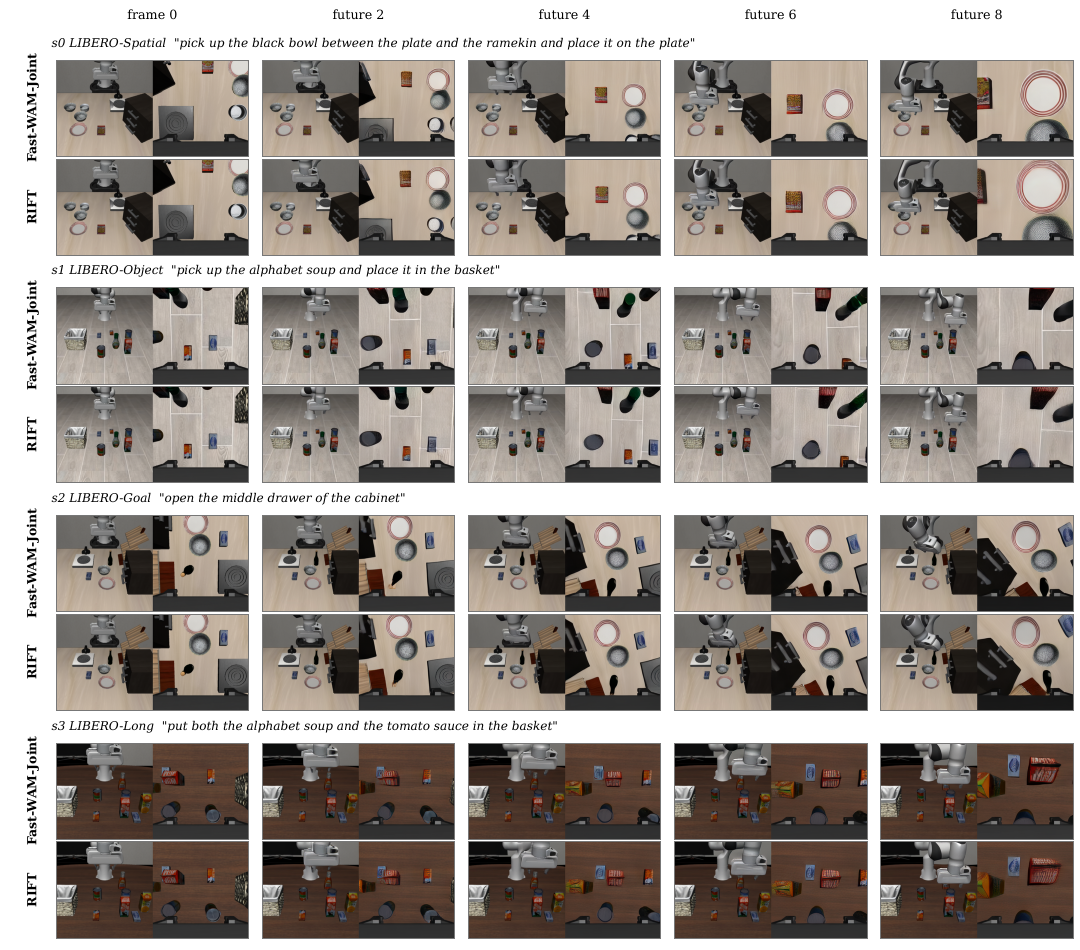}
    \caption{\textbf{Imagined future: Fast-WAM-Joint iterative diffusion versus \method{} one pass.}
    Per state, the top strip is Fast-WAM-Joint's future decoded from its full iterative diffusion
    rollout; the bottom strip is decoded from \method{}'s anticipation tokens after one backbone
    pass. The first column is the shared current observation at frame~0; the remaining columns show
    decoded frames~2, 4, 6, and~8. Visual similarity is illustrative; policy behavior
    is evaluated separately.}
    \label{fig:future_compare}
\end{figure*}

\begin{figure*}[ht]
    \centering
    \includegraphics[width=\textwidth,trim=12.5bp 4bp 2.4bp 3.8bp,clip]{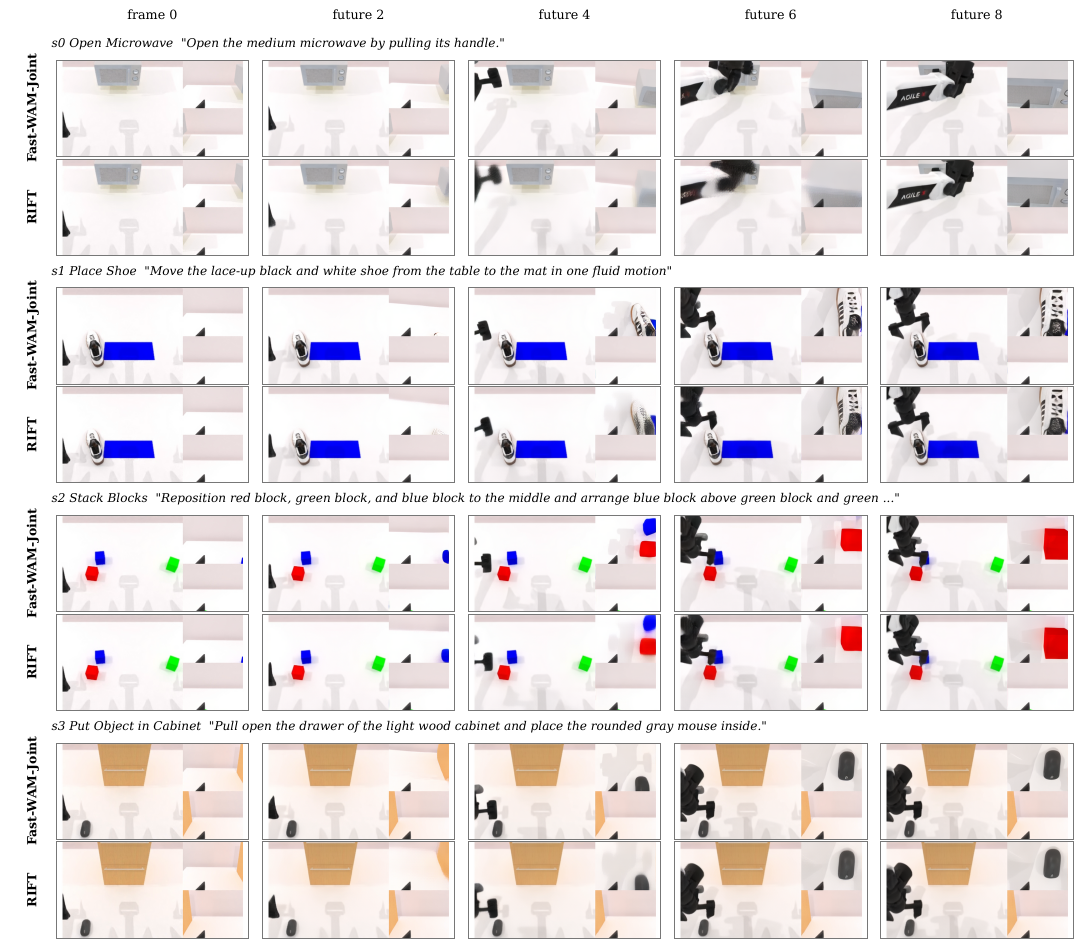}
    \caption{\textbf{RoboTwin~2.0 imagined future: Fast-WAM-Joint iterative diffusion versus \method{}
    one pass.}
    Each row pair starts from the same raw observation from three cameras. The first column is the
    shared observation at frame~0; the remaining columns show decoded frames~2, 4, 6, and~8. The top
    strip uses Fast-WAM-Joint's iterative video diffusion with 20 denoising steps. The bottom strip
    shows \method{}'s anticipation states from one backbone pass, decoded by the trained linear
    diagnostic probe and frozen VAE. The \method{} video decode is diagnostic and is not part of its
    deployed action path; policy behavior is evaluated separately.}
    \label{fig:future_compare_robotwin}
\end{figure*}

\Cref{tab:robotwin_pertask} reports per-task RoboTwin~2.0 results;
\cref{fig:future_compare,fig:future_compare_robotwin} compare matched Fast-WAM-Joint and \method{}
decodes. Visualizations are diagnostic; \cref{sec:finding2} gives closed-loop evidence.
\input{tables/robotwin_pertask_appendix}

%% file: tables/fig2_eeade_ci_appendix.tex
\begin{table}[h!]
\centering
\small
\setlength{\tabcolsep}{3pt}
\caption{\textbf{Uncertainty for the EE-ADE estimates in \cref{fig:future_cache_execution}.}
The table reports percentile bootstrap $95\%$ confidence intervals in centimeters.
Each replicate resamples whole tasks with replacement and recomputes the mean EE-ADE across tasks, keeping
all trials from a sampled task together. The Oracle reference is omitted because its EE-ADE is zero by
definition. Final-clean replay substitutes both final-step keys and values at every
action-denoising step, as in \cref{fig:future_cache_execution}.
All reported intervention estimates use $2{,}000$ paired trials across all $40$ LIBERO tasks.
N/A entries are structural rather
than missing runs: IDM-style models already consume one fixed final-clean future K/V cache, so
final-clean replay coincides with their Oracle reference, and Cosmos-2 exposes only one future
timestep and therefore admits no temporal swap.}
\label{tab:fig2_eeade_ci}
\begin{tabular}{llrc}
\toprule
Model & Intervention & EE-ADE & $95\%$ CI \\
\midrule
Fast-WAM-Joint & Final clean K/V       & 1.908  & $[1.579,\ 2.260]$   \\
      & Spatial shuffle       & 14.306 & $[12.649,\ 16.057]$ \\
      & Temporal swap         & 15.606 & $[13.973,\ 17.189]$ \\
      & Noise                 & 17.221 & $[16.017,\ 18.405]$ \\
      & Frozen present        & 18.255 & $[16.876,\ 19.655]$ \\
      & Masked future         &18.725 & $[17.499,\ 19.953]$ \\
\midrule
Fast-WAM-IDM   & Final clean K/V       & N/A & N/A \\
      & Spatial shuffle       & 15.293 & $[13.214,\ 17.418]$ \\
      & Temporal swap         & 16.325 & $[15.068,\ 17.520]$ \\
      & Noise                 & 17.939 & $[17.070,\ 18.779]$ \\
      & Frozen present        & 21.028 & $[19.901,\ 22.104]$ \\
      & Masked future         &20.378 & $[19.389,\ 21.370]$ \\
\midrule
Cosmos-2
      & Final clean K/V       & 1.690  & $[0.260,\ 3.620]$   \\
      & Spatial shuffle       & 4.976  & $[4.342,\ 5.643]$   \\
      & Temporal swap         & N/A & N/A \\
      & Noise                 & 10.494 & $[8.984,\ 12.034]$  \\
      & Frozen present        & 18.090 & $[16.800,\ 19.323]$ \\
      & Masked future         &11.788 & $[10.516,\ 13.034]$ \\
\midrule
LingBot-VA
      & Final clean K/V       & N/A & N/A \\
      & Spatial shuffle       & 19.820 & $[16.171,\ 22.550]$ \\
      & Temporal swap         & 15.706 & $[12.171,\ 20.044]$ \\
      & Noise                 & 17.466 & $[16.099,\ 18.831]$ \\
      & Frozen present        & 21.323 & $[17.515,\ 25.022]$ \\
      & Masked future         &19.176 & $[17.066,\ 21.131]$ \\
\bottomrule
\end{tabular}
\end{table}

%% file: tables/libero_persuite_appendix.tex
\begin{table*}[!t]
\centering
\caption{\textbf{LIBERO per-suite SR (\%).} Mean\,$\pm$\,std over three seeds ($500$ trials/suite/seed; $2{,}000$ overall). \textbf{Bold}: column bests.}
\label{tab:libero_persuite}
\setlength{\tabcolsep}{4pt}
\renewcommand{\arraystretch}{0.95}
\footnotesize
\newcommand{\sd}[1]{{\scriptstyle\,\pm#1}}
\begin{tabular*}{\textwidth}{@{\extracolsep{\fill}}l*{5}{r}@{}}
\toprule
\textbf{Method} & \textbf{Spatial} & \textbf{Object} & \textbf{Goal} & \textbf{Long} & \textbf{Overall} \\
\midrule
Fast-WAM       & $97.0\sd{0.43}$ & $99.7\sd{0.09}$ & $96.7\sd{0.50}$ & $93.7\sd{0.57}$ & $96.8\sd{0.27}$ \\
PFD            & $98.3\sd{0.19}$ & $99.2\sd{0.33}$ & $98.1\sd{0.62}$ & $93.7\sd{0.41}$ & $97.3\sd{0.12}$ \\
Fast-WAM-Joint & $\mathbf{99.3}\sd{0.41}$ & $99.5\sd{0.25}$ & $98.1\sd{0.62}$ & $96.6\sd{0.57}$ & $98.4\sd{0.26}$ \\
Fast-WAM-IDM   & $\mathbf{99.3}\sd{0.23}$ & $99.3\sd{0.31}$ & $98.1\sd{1.03}$ & $97.5\sd{0.99}$ & $98.6\sd{0.34}$ \\
Cosmos Policy  & $97.8\sd{0.43}$ & $\mathbf{100.0}\sd{0.00}$ & $97.8\sd{0.16}$ & $97.4\sd{0.00}$ & $98.3\sd{0.11}$ \\
LingBot-VA     & $98.5\sd{0.14}$ & $99.6\sd{0.18}$ & $97.2\sd{0.38}$ & $\mathbf{98.5}\sd{0.17}$ & $98.5\sd{0.08}$ \\
\midrule
\textbf{\method{}-L2} & $98.6\sd{0.43}$ & $99.7\sd{0.09}$ & $98.1\sd{0.34}$ & $97.2\sd{0.50}$ & $98.4\sd{0.12}$ \\
\textbf{\method{} (full)}    & $98.7\sd{0.25}$ & $99.7\sd{0.09}$ & $\mathbf{98.8}\sd{0.43}$ & $98.0\sd{0.28}$ & $\mathbf{98.8}\sd{0.17}$ \\
\bottomrule
\end{tabular*}
\end{table*}

%% file: tables/libero_plus_ood_results.tex
\begin{table*}[!t]
\centering
\caption{\textbf{Out-of-distribution robustness on LIBERO-Plus.}
Success rates (\%) use the pinned $10{,}030$-variant protocol with one rollout per variant
(environment/policy seed~42; initial-state index~0) and the benchmark-provided
task-language field. Panels report the official perturbation, difficulty, and source-suite
breakdowns; parenthesized header values are variant counts.
Overall is the variant-micro average rather than an unweighted average of groups. ``Uncl.'' contains
the $121$ variants whose official difficulty label is null. Every populated row is complete over
all $10{,}030$ variants.}
\label{tab:libero_plus_ood}
\setlength{\tabcolsep}{2.5pt}
\renewcommand{\arraystretch}{0.95}
\footnotesize

\textbf{(a) Perturbation category}\par\vspace{2pt}
\begin{tabular*}{\textwidth}{@{\extracolsep{\fill}}l*{8}{r}@{}}
\toprule
\textbf{Method} & \textbf{Camera} & \textbf{Robot} & \textbf{Language} & \textbf{Light} &
\textbf{Texture} & \textbf{Noise} & \textbf{Layout} & \textbf{Overall} \\
& \scriptsize$(1{,}599)$ & \scriptsize$(1{,}550)$ & \scriptsize$(1{,}537)$ &
\scriptsize$(1{,}142)$ & \scriptsize$(1{,}076)$ & \scriptsize$(1{,}601)$ &
\scriptsize$(1{,}525)$ & \scriptsize$(10{,}030)$ \\
\midrule
Fast-WAM                       & 16.51 & 43.42 & 67.79 & 80.12 & 51.58 & 37.85 & 61.18 & 49.73 \\
Fast-WAM-Joint                 & 38.71 & 63.87 & 93.10 & 94.66 & 55.67 & 57.71 & 77.51 & 68.06 \\
Fast-WAM-IDM                   & 46.22 & 69.10 & 94.21 & 92.82 & 56.41 & 61.77 & 81.51 & 71.36 \\
\textbf{\method{} (full)}     & \textbf{59.97} & \textbf{77.48} & \textbf{97.20} & \textbf{96.58} & \textbf{75.09} & \textbf{81.64} & \textbf{82.56} & \textbf{81.07} \\
\bottomrule
\end{tabular*}

\vspace{5pt}
\textbf{(b) Difficulty}\par\vspace{2pt}
\begin{tabular*}{\textwidth}{@{\extracolsep{\fill}}l*{6}{r}@{}}
\toprule
\textbf{Method} & \textbf{Level 1} & \textbf{Level 2} & \textbf{Level 3} & \textbf{Level 4} &
\textbf{Level 5} & \textbf{Uncl.} \\
& \scriptsize$(1{,}644)$ & \scriptsize$(2{,}202)$ & \scriptsize$(2{,}094)$ &
\scriptsize$(1{,}886)$ & \scriptsize$(2{,}083)$ & \scriptsize$(121)$ \\
\midrule
Fast-WAM                       & 68.92 & 57.90 & 50.67 & 44.17 & 28.61 & 74.38 \\
Fast-WAM-Joint                 & 87.59 & 80.02 & 71.59 & 60.50 & 41.43 & \textbf{100.00} \\
Fast-WAM-IDM                   & 89.54 & 83.47 & 73.97 & 64.21 & 46.42 & 99.17 \\
\textbf{\method{} (full)}     & \textbf{94.59} & \textbf{90.87} & \textbf{85.39} & \textbf{77.36} & \textbf{57.99} & 99.17 \\
\bottomrule
\end{tabular*}

\vspace{5pt}
\textbf{(c) Source LIBERO suite}\par\vspace{2pt}
\begin{tabular*}{\textwidth}{@{\extracolsep{\fill}}l*{4}{r}@{}}
\toprule
\textbf{Method} & \textbf{Spatial} & \textbf{Object} & \textbf{Goal} & \textbf{Long} \\
& \scriptsize$(2{,}402)$ & \scriptsize$(2{,}518)$ & \scriptsize$(2{,}591)$ &
\scriptsize$(2{,}519)$ \\
\midrule
Fast-WAM                       & 52.54 & 69.54 & 37.21 & 40.13 \\
Fast-WAM-Joint                 & 81.39 & 72.08 & 56.43 & 63.28 \\
Fast-WAM-IDM                   & 81.72 & 72.40 & 63.80 & 68.20 \\
\textbf{\method{} (full)}     & \textbf{88.72} & \textbf{87.81} & \textbf{73.52} & \textbf{74.79} \\
\bottomrule
\end{tabular*}
\end{table*}

%% file: tables/robotwin_pertask_appendix.tex
\begin{table*}[t]
\centering
\caption{\textbf{Per-task RoboTwin~2.0 success rates (\%) under clean and randomized settings.}
$^{*}$ marks results from \citet{yuan2026fastwam}; bold marks each row's best per setting.}
\label{tab:robotwin_pertask}
\footnotesize
\setlength{\tabcolsep}{2.6pt}
\begin{tabular}{l cc|cc|cc|cc|cc|cc|cc}
\toprule
& \multicolumn{2}{c|}{$\pi_{0.5}^{\,*}$} & \multicolumn{2}{c|}{Motus$^{*}$} & \multicolumn{2}{c|}{LingBot-VA} & \multicolumn{2}{c|}{Fast-WAM-Joint} & \multicolumn{2}{c|}{Fast-WAM} & \multicolumn{2}{c|}{PFD} & \multicolumn{2}{c}{\textbf{\method{} (Ours)}} \\
\cmidrule(lr){2-3}\cmidrule(lr){4-5}\cmidrule(lr){6-7}\cmidrule(lr){8-9}\cmidrule(lr){10-11}\cmidrule(lr){12-13}\cmidrule(lr){14-15}
\textbf{Task} & Clean & Rand. & Clean & Rand. & Clean & Rand. & Clean & Rand. & Clean & Rand. & Clean & Rand. & Clean & Rand. \\
\midrule
Adjust Bottle & \textbf{100} & 99 & 89 & 93 & 90 & 94 & 99 & 98 & \textbf{100} & \textbf{100} & \textbf{100} & \textbf{100} & \textbf{100} & \textbf{100} \\
Beat Block Hammer & 96 & 93 & 95 & 88 & 95 & 98 & \textbf{100} & \textbf{99} & \textbf{100} & 97 & \textbf{100} & 97 & \textbf{100} & 97 \\
Blocks Ranking RGB & 92 & 85 & 99 & 97 & 98 & 98 & \textbf{100} & \textbf{100} & \textbf{100} & 98 & \textbf{100} & 98 & \textbf{100} & 98 \\
Blocks Ranking Size & 49 & 26 & 75 & 63 & \textbf{94} & 96 & 87 & 89 & 93 & \textbf{98} & \textbf{94} & \textbf{98} & \textbf{94} & \textbf{98} \\
Click Alarmclock & 98 & 89 & \textbf{100} & \textbf{100} & 98 & 99 & 99 & \textbf{100} & \textbf{100} & \textbf{100} & \textbf{100} & \textbf{100} & \textbf{100} & \textbf{100} \\
Click Bell & 99 & 66 & \textbf{100} & \textbf{100} & 99 & 99 & \textbf{100} & 97 & \textbf{100} & \textbf{100} & \textbf{100} & \textbf{100} & \textbf{100} & \textbf{100} \\
Dump Bin Bigbin & 92 & \textbf{97} & 95 & 91 & 89 & 96 & 96 & 94 & \textbf{97} & 96 & \textbf{97} & 96 & \textbf{97} & \textbf{97} \\
Grab Roller & \textbf{100} & \textbf{100} & \textbf{100} & \textbf{100} & 99 & 99 & \textbf{100} & \textbf{100} & \textbf{100} & \textbf{100} & \textbf{100} & \textbf{100} & \textbf{100} & \textbf{100} \\
Handover Block & 66 & 57 & 86 & 73 & \textbf{98} & 78 & 92 & \textbf{93} & 97 & 81 & 97 & 82 & 97 & 83 \\
Handover Mic & 98 & 97 & 78 & 63 & 94 & 96 & \textbf{100} & \textbf{100} & 99 & \textbf{100} & 99 & \textbf{100} & 99 & \textbf{100} \\
Hanging Mug & 18 & 17 & 38 & 38 & 40 & 28 & 63 & \textbf{68} & 60 & 62 & 63 & 65 & \textbf{65} & 67 \\
Lift Pot & 96 & 85 & 96 & 99 & 99 & 98 & \textbf{100} & \textbf{100} & \textbf{100} & \textbf{100} & \textbf{100} & \textbf{100} & \textbf{100} & \textbf{100} \\
Move Can Pot & 51 & 55 & 34 & 74 & 94 & 97 & \textbf{98} & \textbf{98} & 90 & 91 & 91 & 92 & 91 & 92 \\
Move Pillbottle Pad & 84 & 61 & 93 & 96 & 98 & \textbf{99} & \textbf{100} & \textbf{99} & \textbf{100} & \textbf{99} & \textbf{100} & \textbf{99} & \textbf{100} & \textbf{99} \\
Move Playingcard Away & 96 & 84 & \textbf{100} & 96 & 99 & 99 & \textbf{100} & \textbf{100} & \textbf{100} & \textbf{100} & \textbf{100} & \textbf{100} & \textbf{100} & \textbf{100} \\
Move Stapler Pad & 56 & 42 & 83 & \textbf{85} & \textbf{91} & 79 & 83 & 84 & 72 & 70 & 74 & 72 & 75 & 74 \\
Open Laptop & 90 & 96 & 95 & 91 & 92 & 94 & 91 & 90 & \textbf{99} & \textbf{100} & \textbf{99} & \textbf{100} & \textbf{99} & \textbf{100} \\
Open Microwave & 34 & 77 & \textbf{95} & \textbf{91} & 82 & 86 & 6 & 25 & 59 & 37 & 62 & 41 & 64 & 45 \\
Pick Diverse Bottles & 81 & 71 & \textbf{90} & \textbf{91} & 89 & 82 & 89 & 85 & 81 & 88 & 82 & 89 & 83 & 89 \\
Pick Dual Bottles & 93 & 63 & 96 & 90 & 99 & 99 & 97 & \textbf{100} & \textbf{100} & 96 & \textbf{100} & 96 & \textbf{100} & 97 \\
Place A2B Left & 87 & 82 & 88 & 79 & 96 & 93 & \textbf{97} & 95 & 94 & 95 & 94 & 95 & 95 & \textbf{96} \\
Place A2B Right & 87 & 84 & 91 & 87 & \textbf{96} & 95 & 94 & 96 & 94 & 96 & 94 & 96 & 95 & \textbf{97} \\
Place Bread Basket & 77 & 64 & 91 & 94 & \textbf{96} & \textbf{95} & 91 & 92 & 93 & 93 & 94 & 94 & 94 & 94 \\
Place Bread Skillet & 85 & 66 & 86 & 83 & 95 & 90 & 88 & 95 & 95 & 95 & 95 & 95 & \textbf{96} & \textbf{96} \\
Place Burger Fries & 94 & 87 & 98 & 98 & 96 & 95 & \textbf{100} & \textbf{99} & 94 & 98 & 94 & 98 & 95 & 98 \\
Place Can Basket & 62 & 62 & \textbf{81} & 76 & \textbf{81} & \textbf{84} & 53 & 35 & 68 & 63 & 70 & 65 & 72 & 68 \\
Place Cans Plasticbox & 94 & 84 & 98 & 94 & 99 & \textbf{99} & 99 & 97 & \textbf{100} & 97 & \textbf{100} & 97 & \textbf{100} & 97 \\
Place Container Plate & \textbf{99} & 95 & 98 & 99 & 98 & 97 & 98 & \textbf{100} & 97 & \textbf{100} & 97 & \textbf{100} & 97 & \textbf{100} \\
Place Dual Shoes & 75 & 75 & 93 & 87 & 94 & 89 & \textbf{95} & 87 & 94 & 89 & 94 & \textbf{90} & \textbf{95} & \textbf{90} \\
Place Empty Cup & \textbf{100} & 99 & 99 & 98 & 99 & 99 & \textbf{100} & \textbf{100} & \textbf{100} & \textbf{100} & \textbf{100} & \textbf{100} & \textbf{100} & \textbf{100} \\
Place Fan & 87 & 85 & 91 & 87 & \textbf{98} & 93 & \textbf{98} & \textbf{97} & 95 & 95 & 95 & 95 & 96 & 96 \\
Place Mouse Pad & 60 & 39 & 66 & 68 & 93 & \textbf{96} & \textbf{94} & 92 & 88 & 89 & 89 & 90 & 89 & 90 \\
Place Object Basket & 80 & 76 & 81 & 87 & \textbf{91} & \textbf{88} & 88 & 80 & 86 & 84 & 87 & 85 & 88 & 86 \\
Place Object Scale & 86 & 80 & 88 & 85 & \textbf{96} & 95 & 95 & \textbf{100} & 93 & 93 & 94 & 94 & 94 & 94 \\
Place Object Stand & 91 & 85 & \textbf{98} & \textbf{97} & \textbf{98} & 96 & 94 & 96 & 89 & 93 & 90 & 94 & 90 & 94 \\
Place Phone Stand & 81 & 81 & 87 & 86 & 96 & 97 & \textbf{100} & \textbf{100} & 99 & 99 & 99 & 99 & 99 & 99 \\
Place Shoe & 92 & 93 & \textbf{99} & 97 & 97 & \textbf{98} & 94 & \textbf{98} & 96 & 97 & 96 & 97 & 96 & 97 \\
Press Stapler & 87 & 83 & 93 & \textbf{98} & 85 & 82 & 52 & 58 & 94 & 96 & 94 & 96 & \textbf{95} & 97 \\
Put Bottles Dustbin & 84 & 79 & 81 & 79 & 87 & 91 & \textbf{95} & \textbf{93} & 91 & 88 & 92 & 89 & 92 & 89 \\
Put Object Cabinet & 80 & 79 & 88 & 71 & 85 & 87 & 93 & \textbf{91} & 93 & 90 & \textbf{94} & \textbf{91} & \textbf{94} & \textbf{91} \\
Rotate QRcode & 89 & 87 & 89 & 73 & \textbf{96} & 91 & 90 & \textbf{94} & 91 & 90 & 92 & 91 & 92 & 91 \\
Scan Object & 72 & 65 & 67 & 66 & \textbf{96} & 91 & 93 & 91 & 90 & 91 & 91 & \textbf{92} & 91 & \textbf{92} \\
Shake Bottle & 99 & 97 & \textbf{100} & 97 & 99 & 97 & \textbf{100} & \textbf{100} & \textbf{100} & \textbf{100} & \textbf{100} & \textbf{100} & \textbf{100} & \textbf{100} \\
Shake Bottle Horizontally & 99 & 99 & \textbf{100} & 98 & 99 & 99 & 99 & \textbf{100} & \textbf{100} & \textbf{100} & \textbf{100} & \textbf{100} & \textbf{100} & \textbf{100} \\
Stack Blocks Three & 91 & 76 & 91 & 95 & 98 & \textbf{98} & \textbf{99} & 96 & 95 & 96 & 95 & 96 & 96 & 97 \\
Stack Blocks Two & 97 & \textbf{100} & \textbf{100} & 98 & 99 & 98 & \textbf{100} & \textbf{100} & \textbf{100} & \textbf{100} & \textbf{100} & \textbf{100} & \textbf{100} & \textbf{100} \\
Stack Bowls Three & 77 & 71 & 79 & \textbf{87} & 86 & 83 & \textbf{87} & 84 & 78 & 81 & 80 & 82 & 81 & 83 \\
Stack Bowls Two & 95 & 96 & \textbf{98} & \textbf{98} & 94 & \textbf{98} & 96 & 97 & 93 & 94 & 94 & 94 & 94 & 95 \\
Stamp Seal & 79 & 55 & 93 & 92 & 96 & 97 & \textbf{97} & \textbf{98} & 89 & 97 & 90 & 97 & 90 & 97 \\
Turn Switch & 62 & 54 & \textbf{84} & \textbf{78} & 44 & 45 & 71 & 75 & 60 & 66 & 63 & 68 & 65 & 70 \\
\midrule
\textbf{Average} & 82.7 & 76.8 & 88.7 & 87.0 & 92.4 & 91.4 & 91.0 & 91.1 & 91.9 & 91.6 & 92.5 & 92.1 & \textbf{92.9} & \textbf{92.6} \\
\bottomrule
\end{tabular}
\end{table*}